\documentclass[letterpaper]{article} % DO NOT CHANGE THIS
\usepackage{aaai2027}  % DO NOT CHANGE THIS
\usepackage[hyphens]{url}  % DO NOT CHANGE THIS
\usepackage{graphicx} % DO NOT CHANGE THIS
\usepackage{natbib}  % DO NOT CHANGE THIS AND DO NOT ADD ANY OPTIONS TO IT
\usepackage{caption} % DO NOT CHANGE THIS AND DO NOT ADD ANY OPTIONS TO IT
\usepackage{algorithm}
\usepackage{algorithmic}

\usepackage{newfloat}
\usepackage{listings}
\DeclareCaptionStyle{ruled}{labelfont=normalfont,labelsep=colon,strut=off} % DO NOT CHANGE THIS
\floatstyle{ruled}
\newfloat{listing}{tb}{lst}{}
\floatname{listing}{Listing}

\usepackage{booktabs}

\usepackage{booktabs}
\usepackage{tabularx}
\usepackage{amsmath}
\usepackage{multirow}
\usepackage{array}
\usepackage[table]{xcolor}
\usepackage{amssymb} % 必须引入，用于支持对勾符号
\usepackage{threeparttable}
\usepackage{arydshln} % 必须放在最后

\title{Verification-Notebook Learning for Source-Aware Multimodal Misinformation Detection}
\author{
    Junyuan Tan
}
\affiliations{
    Independent Researcher\\
    junyuan.tan.cs@gmail.com
}

\begin{document}

\nocopyright
\maketitle

\begin{abstract} Multimodal misinformation verification is challenging because misleading signals may come from different parts of a post and require different forms of evidence. LVLMs are well suited to this task, but their verification performance often depends on the inference procedure applied to each instance. Existing methods improve this procedure through stronger prompting, retrieval, or deliberation, but rarely retain the verification patterns learned from previous examples. We propose Verification-Notebook Learning (VNL), a non-parametric framework that learns an external verification procedure for a frozen LVLM before inference. VNL builds a compact notebook of decision principles, evidence cues, and recurring pitfalls from prior verification experience. The notebook remains fixed during inference and guides the verification of new examples. Rather than updating model parameters or storing demonstrations, VNL records learned knowledge in an artifact that can be inspected directly. Experiments show that VNL consistently outperforms a range of competitive baselines. Further analyses show that the Verification Notebook improves fine-grained source attribution while remaining compact and interpretable, providing an effective way to accumulate verification knowledge without model training. \end{abstract}

\section{Introduction}

The growing diversity of multimodal misinformation makes source-aware verification increasingly important, as misleading content may come from the text, the image, or the relation between them \cite{liu2025mmfakebench}. A news post may be deceptive because its textual claim is false \cite{thorne2018fever,shu2020fakenewsnet}, its visual content has been manipulated \cite{shao2023detecting}, or two otherwise plausible modalities have been paired to imply a false association \cite{luo2021newsclippings,abdelnabi2022open}. Detecting such cases requires more than a binary judgment of veracity. A reliable system must identify the source of the distortion, determine which evidence to examine, and weigh textual, visual, cross-modal, and external signals when they conflict \cite{liu2025mmfakebench}.

LVLMs provide a strong foundation for this setting \cite{liu2023visual,li2023blip}. They can describe visual content, interpret textual claims, compare information across modalities, and reason with external knowledge. However, their verification behavior remains brittle \cite{bai2024hallucination}. Under direct prompting, an LVLM may conflate visual plausibility with factual correctness, interpret weak semantic relatedness as cross-modal consistency, or attribute uncertainty to visual manipulation without sufficient evidence. These errors are not solely failures of perception or retrieval. They expose a more fundamental limitation: current LVLM-based verification lacks a stable and reusable procedure for determining which evidence to examine and how to prioritize conflicting signals.

Most recent LVLM-based systems address this limitation with more elaborate inference procedures. They augment each instance with retrieved evidence or tool outputs \cite{cui2026t2agent}, decompose verification into multiple reasoning steps \cite{liu2025mmfakebench}, or use questions, search, and debate to support agentic deliberation \cite{beigi2025can,xuan2024lemma,liu2025truth}. These systems seek to improve inference-time reasoning while leaving the underlying model unchanged. Although they can improve individual predictions, they do not accumulate verification knowledge across examples. Each test case starts a new reasoning process, and useful patterns discovered during inference remain temporary. The system may therefore perform more reasoning at inference time without learning a stable verification procedure before deployment.

Earlier supervised approaches provide a different way to learn from task experience. Fine-tuned multimodal detectors, representation-level classifiers, and parameter-efficient adaptation methods can encode this experience in model parameters when labeled data and trainable backbones are available \cite{wang2018eann,ying2023bootstrapping,qi2024sniffer,yan2025trust}. These methods show that development experience can improve multimodal verification. However, they differ from the LVLM-based inference systems considered in this work. Parameter learning requires access to the model, training infrastructure, and careful optimization, while the resulting knowledge is difficult to inspect or transfer. We do not seek to replace parametric adaptation. Instead, we study whether development-set experience can improve black-box LVLM verification through an external, auditable, and deployment-friendly form of learning.

We investigate whether a frozen LVLM can use prior verification experience through an explicit external procedure rather than relying solely on instance-level inference traces or parameter updates. Source-aware misinformation detection needs a learning mechanism that produces neither a temporary reasoning trace nor an updated set of parameters, but an external verification procedure. This procedure should be learned from development experience, remain fixed during inference, and be explicit enough to support analysis and revision.

We propose Verification-Notebook Learning, a two-stage non-parametric framework that learns reusable verification knowledge outside the LVLM. Its output is a Verification Notebook, a compact and structured collection of source-aware notes covering decision guidance, mistakes to avoid, and observable evidence cues. The notebook is not a single instruction template of the kind used in prompting, nor does it store individual cases for later reuse. Instead, it summarizes verification experience as general rules and evidence cues that guide future reasoning.

During the learning stage, a Verifier-Editor loop constructs the notebook from labeled development data. The Verifier consults the current notebook and produces a verdict together with a reasoning trace. The Editor compares batches of verification records with the available supervision, checks whether the relevant evidence was examined, identifies where the reasoning succeeded or failed, and converts reusable insights into concise notebook updates. Each update is assigned to the appropriate notebook function, either decision guidance and mistakes to avoid or observable evidence cues. Redundant and case-specific notes are discarded. Candidate notebooks are selected according to validation performance. The LVLM parameters remain frozen throughout this process, and learning occurs only through updates to the external notebook. During inference, the selected Verification Notebook remains fixed and guides source-aware verification of unseen examples. It may use the same type of external evidence available to inference-only baselines. Test examples do not update the notebook, and no feedback from the test distribution is used.

Verification-Notebook Learning offers an alternative to both increasingly expensive inference and opaque parametric adaptation. Unlike inference-only agentic reasoning, it converts repeated verification experience into persistent knowledge. Unlike fine-tuning, it is lightweight, auditable, and compatible with black-box models. We hypothesize that robust multimodal verification requires both strong test-time reasoning and reusable guidance about what to verify, when to verify it, and how to prioritize different forms of evidence.

Our contributions are as follows:
\begin{itemize}
    \item We propose Verification-Notebook Learning as a non-parametric learning paradigm for source-aware multimodal misinformation detection.
    \item We introduce a Verifier-Editor mechanism that distills labeled verification experience into a fixed, source-aware Verification Notebook.
    \item We show that Verification-Notebook Learning improves multimodal misinformation detection, while analyses of notebook evolution and decision-boundary behavior clarify how explicit verification knowledge affects model decisions.
\end{itemize}

\begin{figure*}[t]
    \centering
\includegraphics[
  width=\textwidth,
  trim=4pt 4pt 4pt 4pt,  % 左 下 右 上，按实际微调
  clip
]{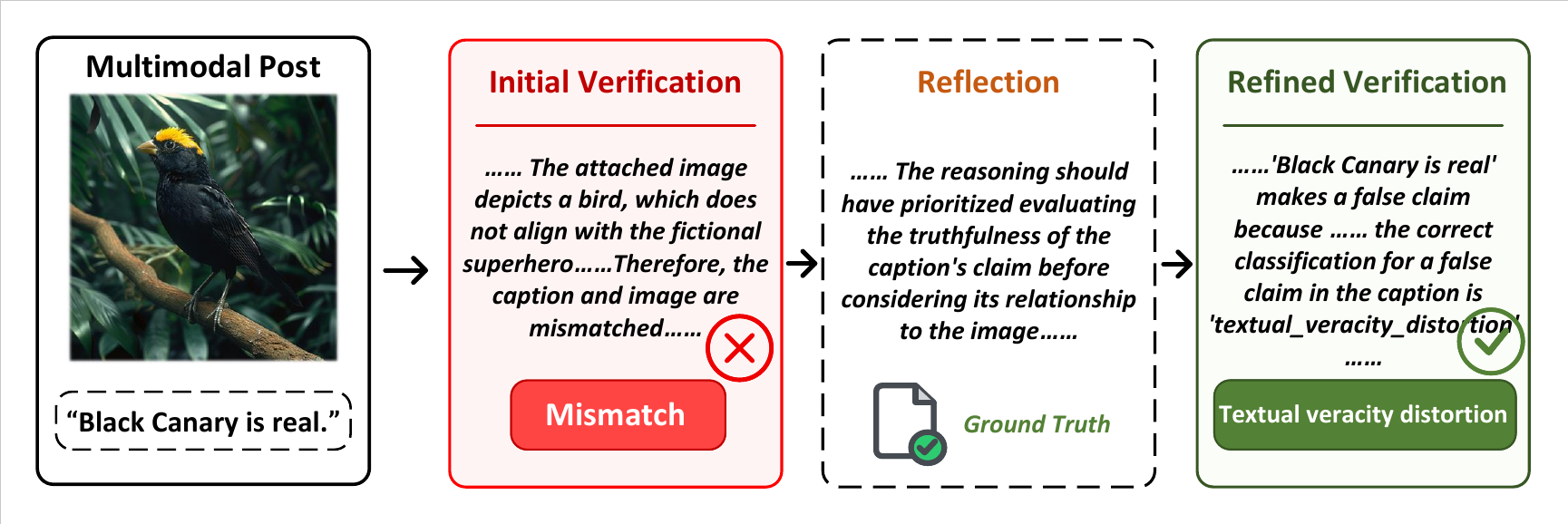}
    \caption{A motivating example of fine-grained source attribution.}
    \label{fig:pre_case}
\end{figure*}

\section{Related Work}

\subsection{Multimodal Misinformation Detection}

Multimodal misinformation detection has often been studied through supervised representation learning over heterogeneous signals. Before LVLMs became widely used for this task, most systems trained task-specific detectors using textual content, visual content, social context, event information, or propagation patterns \cite{wang2018eann,singhal2019spotfake,shu2019defend,bian2020rumor}. This research includes multi-view and multimodal representation learning \cite{ying2023bootstrapping,wu2023see}, cross-modal fusion \cite{chen2022cross}, attention-based alignment \cite{qian2021hierarchical}, and graph- or propagation-aware detection \cite{wang2020fake}. Although these methods differ in architecture and supervision, they all combine multiple forms of evidence rather than treating text and images as independent inputs.

The task has also evolved from coarse veracity classification to fine-grained source analysis. Early studies focused primarily on binary fake-news detection \cite{nakamura2020fakeddit}, whereas later work examined different sources of misleading information \cite{liu2025mmfakebench}. A post may be deceptive because its textual claim is false \cite{thorne2018fever}, its image has been manipulated or synthetically generated \cite{shao2023detecting}, or an otherwise plausible image and caption have been paired out of context \cite{luo2021newsclippings,aneja2021cosmos}. Evidence-grounded multimodal fact-checking also shows that some claims require information beyond the post itself \cite{yao2023end}. Recent source-aware benchmarks make these distinctions explicit by requiring the model to determine whether the distortion comes from the text, the image, or the cross-modal relation \cite{liu2025mmfakebench}. This formulation is more demanding than binary classification because the system must both detect misinformation and attribute the misleading signal to the correct source.

Our work adopts this source-aware setting and studies a different way to use development-set experience for multimodal misinformation detection. Existing learning-based methods typically encode such experience in model parameters or learned representations. In contrast, VNL keeps the LVLM fixed and records task experience in an external Verification Notebook. This design makes verification knowledge explicit, reusable, and easy to inspect.

\subsection{LVLM-based Multimodal Verification}

LVLMs provide a different approach to multimodal misinformation detection. Rather than training a task-specific classifier, recent systems often formulate verification as prompted multimodal reasoning. The model interprets the caption, examines the image, compares textual and visual signals, and produces a source-aware decision with an explanation \cite{zhang2023towards,tahmasebi2024multimodal}. A single LVLM can therefore support visual description, textual reasoning, cross-modal comparison, and knowledge-based inference. However, verification quality depends strongly on the protocol provided to the model, including label definitions, evidence-checking instructions, question design, and reasoning format \cite{xuan2024lemma,xiao2025xfacta,beigi2025can}.

A common research direction is to improve LVLM verification through stronger test-time reasoning. Some methods provide retrieved evidence or entity-level context to ground factual claims \cite{xuan2024lemma,li2025cmie}. Others decompose verification into intermediate questions, fixed workflows, or tool-assisted reasoning steps \cite{liu2025mmfakebench,xuan2024lemma,xu2025mdam3}. More agentic systems allocate additional test-time computation through planning, adaptive reasoning, multiple agents, or debate-style deliberation \cite{cui2026t2agent,liu2025truth}. These studies show that stronger reasoning and richer context can improve LVLM-based verification. However, they primarily change how each test instance is processed. The verifier typically starts a new reasoning process for every example, and useful verification patterns discovered during inference are not retained as a fixed task procedure.

Source-aware multimodal misinformation detection requires the model to examine and prioritize textual veracity, visual authenticity, and cross-modal consistency. VNL addresses this challenge by using labeled verification records from the development set to build a compact notebook before deployment, rather than retrieving similar cases or introducing additional test-time agents. The notebook records source-attribution rules, evidence cues, and recurring mistakes that help distinguish textual distortion, visual distortion, and mismatch. It remains fixed throughout test-time inference.

\begin{figure}[t]
\centering
\includegraphics[width=\columnwidth]{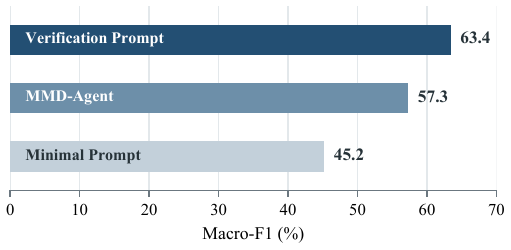}
\caption{Preliminary macro-F1 comparison of direct LVLM prompting and MMD-Agent on the four-way source-attribution task.}
\label{fig:pre1}
\end{figure}

\section{Preliminaries}

\subsection{Problem Formulation}

We study source-aware multimodal misinformation detection. Each example is a news post $x=(c,I,e)$, where $c$ is a textual claim or caption, $I$ is the associated image, and $e$ is optional external evidence. The task is to predict a label $y\in\mathcal{Y}$ from a source-aware label space. In the multiclass setting, $\mathcal{Y}$ contains four labels: original content, textual veracity distortion, visual veracity distortion, and mismatch. The model must determine whether a post is misleading and whether the distortion comes from the text, the image, or the relation between them.

Given a frozen LVLM $f_\theta$, a development set $\mathcal{D}_{\mathrm{dev}}=\{(x_i,y_i)\}_{i=1}^{n}$, and a held-out validation set $\mathcal{D}_{\mathrm{val}}$, we aim to learn an external verification procedure without updating the parameters of $f_\theta$. We represent this procedure as a Verification Notebook $\mathcal{N}$, which is learned from the development set and kept fixed during inference.

\subsection{Motivating Observations}

\paragraph{Direct LVLM verification can be substantially improved.} Recent LVLM-based methods for misinformation detection often use multi-step or multi-agent reasoning to improve performance. Our preliminary results show that direct LVLM verification also benefits from careful prompting, as illustrated in Figure~\ref{fig:pre1}. A simple zero-shot prompt achieves an F1 score of 45.2, while a verification-oriented prompt with clearer task instructions and source-aware label definitions raises the score to 63.4. In comparison, MMD-Agent achieves an F1 score of 57.3. This comparison suggests that precise guidance can substantially improve direct verification.

\begin{figure*}[t]
    \centering
\includegraphics[
  width=\textwidth,
  trim=4pt 4pt 4pt 4pt,  % 左 下 右 上，按实际微调
  clip
]{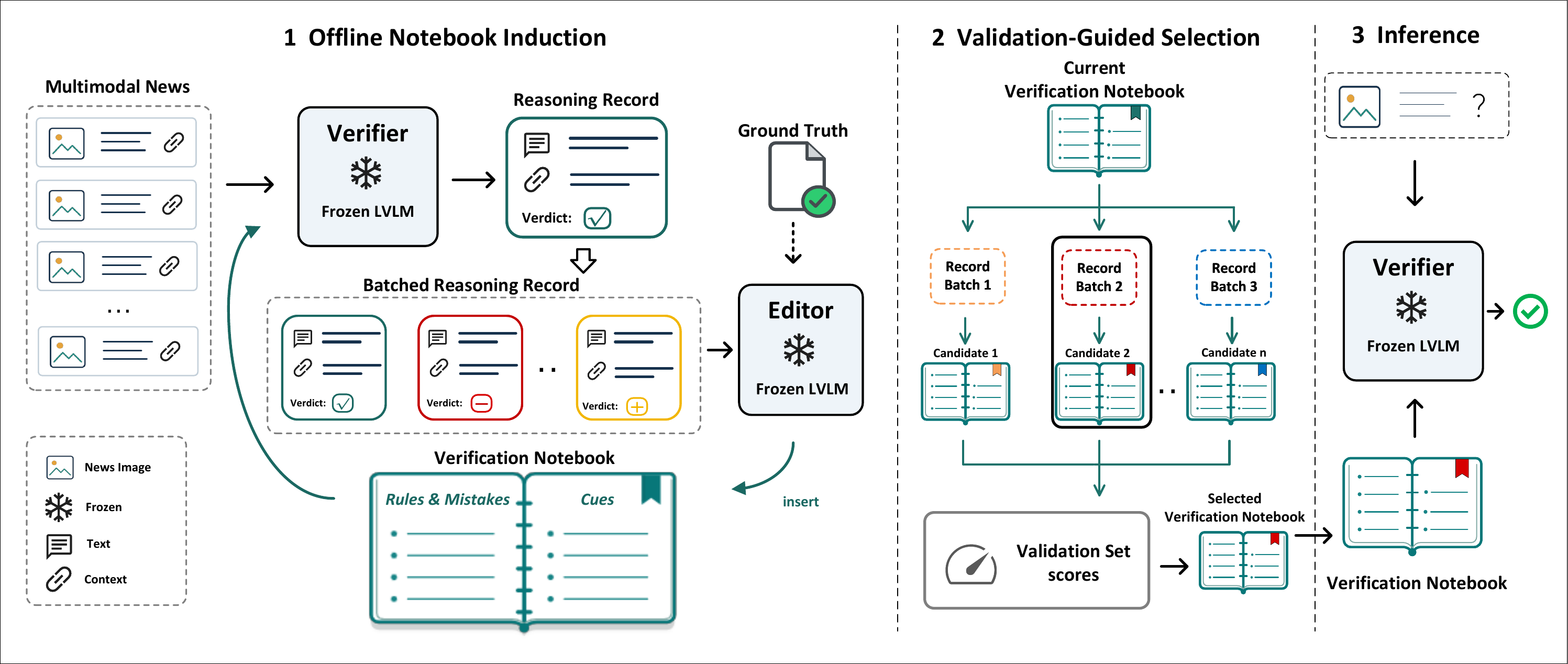}
    \caption{Overview of Verification-Notebook Learning (VNL).}
    \label{fig:vnl_overview}
\end{figure*}

\paragraph{Source attribution requires fine-grained distinctions among labels.} Source-aware misinformation detection requires careful distinctions among closely related labels. Figure~\ref{fig:pre_case} shows an example in which the caption claims that ``Black Canary is real.'' The initial verifier correctly recognizes Black Canary as a fictional character but predicts mismatch because it focuses on the inconsistency between the caption and the associated image. After reflection, it revises the prediction to textual veracity distortion because the caption itself contains a false factual claim.

Detecting that a post is misleading is not sufficient to identify the source of the distortion. Without reusable guidance, an inference-only method must resolve these fine-grained label distinctions independently for each new example and cannot retain useful conclusions from previous reasoning. In this case, reflection yields a general rule: when a caption contains a false factual claim, the factual accuracy of the caption should be evaluated before the post is classified as a mismatch based on image-caption alignment. Retaining such rules allows the verifier to apply them to later examples instead of deriving them again during each inference.

\section{Method}
Figure~\ref{fig:vnl_overview} provides an overview of VNL. 

\subsection{Verification Notebook}

A Verification Notebook $\mathcal{N}$ is a compact textual record of reusable verification knowledge rather than a collection of instance-level demonstrations. Each note describes a general decision rule, an evidence cue, a label-boundary condition, or a recurring reasoning failure. In our implementation, the notes are organized into two functional sections: Decision Rules \& Mistakes to Avoid and Observable Evidence Cues. The former contains decision procedures, label-boundary logic, arbitration principles, and recurring reasoning pitfalls, whereas the latter records concrete textual, visual, and cross-modal evidence cues. This structure keeps the notebook procedural, evidence-centered, and applicable across instances.

\subsection{Verifier-Editor Learning}

Verification-Notebook Learning uses two prompted LVLM operators: a Verifier and an Editor, denoted by $V_\theta$ and $E_\theta$, respectively. Each operator is implemented by applying a role-specific prompt to a frozen LVLM. The shared subscript $\theta$ indicates that the model parameters remain fixed and that learning occurs only through updates to the external notebook.

At update step $t$, the Verifier receives an instance $x_i$ and the current notebook $\mathcal{N}_t$. It produces a reasoning trace $r_i$ and a prediction $\hat{y}_i$:
\begin{equation}
(r_i,\hat{y}_i)=V_\theta(x_i,\mathcal{N}_t), \quad \hat{y}_i\in\mathcal{Y}.
\end{equation}
The Verifier uses $\mathcal{N}_t$ as accumulated verification experience and consults the entries relevant to the current instance. During notebook induction, the system records the reasoning trace, prediction, verification outcome, and reference label:
\begin{equation}
z_i=(x_i,r_i,\hat{y}_i,y_i,\mathbb{I}[\hat{y}_i=y_i]).
\end{equation}

For a mini-batch $B$, the Editor receives the current notebook, a batch of verification records, and the corresponding textual case contexts:
\begin{equation}
o_t=E_\theta(\mathcal{N}_t,\{z_i\}_{i\in B},\{q_i\}_{i\in B}),
\end{equation}
where $q_i$ denotes the textual context provided to the editing stage. Using the available supervision signal, the Editor determines whether the Verifier examined the relevant evidence, identifies where the reasoning succeeded or failed, and extracts insights that generalize beyond the current example. It then proposes notebook operations $o_t$ that add concise, reusable, and non-redundant notes:
\begin{equation}
\mathcal{N}_{t+1}=\mathrm{Apply}(\mathcal{N}_t,o_t).
\end{equation}
Each new note is assigned to exactly one notebook function: decision guidance and mistakes to avoid, or observable evidence cues. In the batched induction pipeline, the Editor processes textual representations of the verification records and textual case contexts rather than raw image inputs. Visual evidence informs notebook updates only when it is described in the reasoning trace of the Verifier or in the textual case context. Highly similar notes are filtered to keep the notebook compact and reduce redundancy.

\subsection{Offline Notebook Induction}

We learn the notebook offline from $\mathcal{D}_{\mathrm{dev}}$. For each mini-batch $B\subset\mathcal{D}_{\mathrm{dev}}$, all instances are processed independently using the same notebook snapshot. The resulting verification records are accumulated, and the Editor periodically updates the notebook based on these records. Batch-level updates reduce sensitivity to individual instances and help the notebook capture recurring verification patterns.

At fixed selection intervals, candidate notebooks are evaluated on $\mathcal{D}_{\mathrm{val}}$. Let $S_{\mathrm{val}}(\mathcal{N};\mathcal{D}_{\mathrm{val}})$
denote the held-out selection score obtained when the Verifier uses notebook $\mathcal{N}$. The selected notebook is
\begin{equation}
\mathcal{N}^{*}=\arg\max_{\mathcal{N}_t} S_{\mathrm{val}}(\mathcal{N}_t;\mathcal{D}_{\mathrm{val}}).
\end{equation}
In our experiments, $S_{\mathrm{val}}$ is the primary validation metric for the task.

\subsection{Validation-Guided Multi-Branch Selection}

We also use a validation-guided multi-branch variant corresponding to best-of-$N$ notebook induction. At the beginning of selection period $p$, $N$ candidate notebooks are forked from the current best notebook. Each branch follows the same Verifier-Editor update procedure but uses independently sampled development mini-batches. Editing is performed at a fixed frequency, and held-out evaluation occurs only at the end of each period.

Let $\mathcal{N}_{p,j}$ denote branch $j$ in period $p$. The candidate for the period is selected by
\begin{equation}
j^{*}=\arg\max_{j\in\{1,\ldots,N\}} S_{\mathrm{val}}(\mathcal{N}_{p,j};\mathcal{D}_{\mathrm{val}}).
\end{equation}
The selected branch replaces the current best notebook if its held-out score matches or exceeds the current best score. Otherwise, the previous best notebook is retained. This procedure searches over discrete notebook updates while keeping the LVLM parameters fixed. Algorithm~\ref{alg:vnl} summarizes the validation-guided multi-branch procedure used to induce and select the Verification Notebook.

\subsection{Inference}

After induction, the selected notebook $\mathcal{N}^{*}$ remains fixed. For an unseen instance $x$, inference is performed as
\begin{equation}
(r,\hat{y})=V_\theta(x,\mathcal{N}^{*}).
\end{equation}
The notebook provides learned procedural guidance to the Verifier, while the underlying LVLM remains unchanged.

\begin{algorithm}[tb]
\caption{Validation-Guided Multi-Branch Verification-Notebook Learning}
\label{alg:vnl}
\textbf{Input}: Development set $\mathcal{D}_{\mathrm{dev}}$, validation set $\mathcal{D}_{\mathrm{val}}$, textual case contexts $\{q_i\}$, frozen Verifier $V_\theta$, frozen Editor $E_\theta$, and initial notebook $\mathcal{N}_0$\\
\textbf{Parameter}: Number of selection periods $P$, number of branches $N$, update steps per period $K$, mini-batch size $b$, and editing interval $h$ satisfying $h\mid K$\\
\textbf{Output}: Selected Verification Notebook $\mathcal{N}^{*}$
\begin{algorithmic}[1]
\STATE $\mathcal{N}^{*}\leftarrow\mathcal{N}_0$
\STATE $s^{*}\leftarrow S_{\mathrm{val}}(\mathcal{N}^{*};\mathcal{D}_{\mathrm{val}})$
\FOR{$p=1$ to $P$}
    \FOR{$j=1$ to $N$}
        \STATE $\mathcal{N}_{p,j}\leftarrow\mathcal{N}^{*}$
        \STATE $\mathcal{Z}_{j}\leftarrow\varnothing$, $\mathcal{Q}_{j}\leftarrow\varnothing$
        \FOR{$k=1$ to $K$}
            \STATE Sample a mini-batch $B_{p,j,k}\subset\mathcal{D}_{\mathrm{dev}}$ of size $b$
            \FOR{each $(x_i,y_i)\in B_{p,j,k}$}
                \STATE $(r_i,\hat{y}_i)\leftarrow V_\theta(x_i,\mathcal{N}_{p,j})$
                \STATE $z_i\leftarrow(x_i,r_i,\hat{y}_i,y_i,\mathbb{I}[\hat{y}_i=y_i])$
                \STATE $\mathcal{Z}_{j}\leftarrow\mathcal{Z}_{j}\cup\{z_i\}$
                \STATE $\mathcal{Q}_{j}\leftarrow\mathcal{Q}_{j}\cup\{q_i\}$
            \ENDFOR
            \IF{$k\bmod h=0$}
                \STATE $o\leftarrow E_\theta(\mathcal{N}_{p,j},\mathcal{Z}_{j},\mathcal{Q}_{j})$
                \STATE $\mathcal{N}_{p,j}\leftarrow\mathrm{Apply}(\mathcal{N}_{p,j},o)$
                \STATE $\mathcal{Z}_{j}\leftarrow\varnothing$, $\mathcal{Q}_{j}\leftarrow\varnothing$
            \ENDIF
        \ENDFOR
        \STATE $s_{p,j}\leftarrow S_{\mathrm{val}}(\mathcal{N}_{p,j};\mathcal{D}_{\mathrm{val}})$
    \ENDFOR
    \STATE $j^{*}\leftarrow\arg\max_{j\in\{1,\ldots,N\}}s_{p,j}$
    \IF{$s_{p,j^{*}}\geq s^{*}$}
        \STATE $\mathcal{N}^{*}\leftarrow\mathcal{N}_{p,j^{*}}$
        \STATE $s^{*}\leftarrow s_{p,j^{*}}$
    \ENDIF
\ENDFOR
\STATE \textbf{return} $\mathcal{N}^{*}$
\end{algorithmic}
\end{algorithm}

\section{Experiments}

\subsection{Experimental Setup}

\paragraph{Dataset and splits.}
We evaluate VNL on MMFakeBench~\cite{liu2025mmfakebench} using its four-way source-attribution setting. Each instance contains a caption and an associated image, and the model predicts one of four labels: original, textual veracity distortion, visual veracity distortion, or mismatch. The model must determine both whether the post is misleading and whether the distortion originates from the text, the image, or the cross-modal relation.

We adopt a learn-then-deploy protocol. We split the original MMFakeBench validation set into a development set and a held-out validation set at a ratio of 7:3, yielding 700 and 300 examples, respectively. The development set is used for notebook induction, whereas the held-out validation set is used for notebook selection. For final evaluation, we sample 2,000 examples from the original MMFakeBench test set while preserving the class distribution. The resulting test set contains 600 original examples, 600 textual-distortion examples, 200 visual-distortion examples, and 600 mismatch examples. Neither the test examples nor their labels are used for notebook induction, notebook selection, or prompt development.

\paragraph{Evaluation metrics.}
We report macro-averaged precision, recall, and F1, together with accuracy. Macro-F1 is the primary metric because the task is class-imbalanced, with substantially fewer visual-distortion examples than examples from the other classes. Compared with accuracy, macro-F1 better reflects performance across different sources of misinformation. Candidate notebooks are selected based on macro-F1 on the held-out validation set.

\paragraph{Implementation details.}
All LVLM operators in VNL are implemented using GPT-4o with role-specific prompts, and the model parameters remain frozen throughout the experiments. VNL obtains Wikipedia evidence using the evidence-acquisition pipeline of MMD-Agent \cite{liu2025mmfakebench} and provides it in the input context. Offline notebook induction uses a mini-batch size of 20. The Editor updates the notebook after each mini-batch, and candidate notebooks are evaluated on the held-out validation set every 40 examples. We use bge-base-en-v1.5 to detect semantic duplicates and filter notes with a similarity score above 0.9. For validation-guided notebook search, the first selection period uses eight branches to broaden the initial search, whereas each subsequent period uses four branches. At the end of each period, the branch with the highest validation macro-F1 is selected and accepted only if its score matches or exceeds the best validation score observed so far. Notebook induction stops if the best validation macro-F1 does not improve for two consecutive selection periods. The final notebook is fixed before test-time inference. Because notebook updates depend on the order of the development examples, we run the complete learning and evaluation pipeline with three random seeds and report the mean performance.

\subsection{Baselines}

We compare VNL with several baselines. Unless otherwise specified, baselines implemented in our experiments use the same GPT-4o model, label space, data split, and test instances as VNL.

\paragraph{Standard Prompt (SP).}
SP follows the LVLM evaluation protocol reported in MMFakeBench~\cite{liu2025mmfakebench}. It prompts the model to classify a caption-image pair into the benchmark label space without external memory or learned verification knowledge.

\paragraph{MMD-Agent.}
MMD-Agent is the agentic multimodal misinformation detection system introduced with MMFakeBench~\cite{liu2025mmfakebench}. It retrieves entity-centered evidence from Wikipedia and performs multi-stage reasoning over textual, visual, and external information for each instance.

\paragraph{Direct.}
Direct is a zero-shot LVLM baseline that receives a caption and an image and predicts one of the four source-aware labels without demonstrations, retrieved cases, external evidence, or notebook guidance.

\paragraph{Chain-of-Thought (CoT).}
CoT adds a generic chain-of-thought instruction to Direct, asking the model to reason step by step before producing the final label.

\paragraph{Wiki Context.}
Wiki Context augments the input with Wikipedia search results generated by the evidence-acquisition stage used in MMD-Agent. The model receives the retrieved textual evidence together with the caption and image.

\paragraph{ICL.}
In-Context Learning (ICL) \cite{brown2020language} uses a fixed, class-balanced set of four labeled demonstrations, with one example from each class. The same set of demonstrations is used for every test instance.

\paragraph{CR.}
Case Retrieval (CR) constructs a case bank from the development set and retrieves the three most similar labeled cases for each query based on caption similarity. The retrieved captions, images, and labels are provided to the LVLM.

\paragraph{ICL + Wiki and CR + Wiki.}
ICL + Wiki combines the fixed, class-balanced demonstrations with Wikipedia evidence. CR + Wiki combines the three retrieved cases with Wikipedia evidence.

For the controlled baselines, we use two prompt protocols. The Minimal Prompt specifies the input format and candidate labels while providing limited task guidance. The Verification Prompt explicitly defines the four source-aware labels and instructs the LVLM to examine textual veracity, image authenticity, and cross-modal consistency separately.

\subsection{Main Results}

Table~\ref{tab:results} reports the results on the four-way source-attribution task. MMD-Agent substantially improves on the standard prompting protocol, confirming the value of multi-stage reasoning and external evidence for multimodal misinformation detection. However, Direct with the Verification Prompt already outperforms MMD-Agent using a single LVLM call. This result indicates that a complex inference procedure cannot fully compensate for an under-specified task interface. Detailed task instructions and clear label definitions alone provide strong guidance for source-aware LVLM verification.

The comparison between the Minimal Prompt and the Verification Prompt further demonstrates the importance of task specification. The Verification Prompt consistently achieves stronger results across inference settings, showing that LVLMs benefit from detailed verification instructions and clear source-attribution label definitions. In contrast, generic chain-of-thought prompting does not provide a consistent improvement. CoT slightly outperforms Direct under the Minimal Prompt but performs worse than Direct under the Verification Prompt. Longer reasoning traces therefore do not necessarily improve the decision criteria required for fine-grained source attribution.

External evidence and development examples also improve performance, although the gains remain limited when this information is supplied directly in the prompt. Under the Verification Prompt, Wiki Context improves on Direct, indicating that retrieved factual evidence helps resolve some entity-level and textual-veracity uncertainty. Example-based methods such as ICL and CR provide additional information about label semantics and decision boundaries, and combining case retrieval with external evidence yields the strongest prompt-based use of development information. Nevertheless, these methods primarily use development examples as instance-level references during inference, leaving their reusable verification knowledge implicit.

VNL achieves the best overall performance, with a macro-F1 of 73.2\% and an accuracy of 71.8\%. Compared with CR + Wiki, which also uses retrieved evidence and development-set cases, VNL improves macro-F1 by 5.4 percentage points and accuracy by 4.9 percentage points. Compared with MMD-Agent, VNL improves macro-F1 by 15.9 percentage points and accuracy by 16.0 percentage points. These gains show that the Verification Notebook uses development experience more effectively than methods that directly retrieve or insert individual cases. Rather than storing examples as demonstrations, VNL distills them into source-aware decision rules, observable evidence cues, label-boundary conditions, and recurring mistakes to avoid. These notes guide the Verifier in determining which evidence to examine, how to prioritize textual, visual, and cross-modal signals, and when to attribute misinformation to textual distortion rather than mismatch or visual distortion. Because the notebook is learned offline and remains fixed during inference, VNL converts development-set experience into explicit and reusable verification knowledge while preserving the black-box LVLM setting.

\begin{table}[t]
\centering
\caption{Results on the four-way source-attribution task of MMFakeBench.
F1 is the primary metric. P, R, and F1 are macro-averaged. Bold denotes
the best overall result, and underline denotes the strongest inference-only
baseline. Results of our implementations are averaged over three runs.}
\label{tab:results}
\small
\setlength{\tabcolsep}{4pt}
\begin{tabularx}{\columnwidth}{
    >{\raggedright\arraybackslash}X
    >{\centering\arraybackslash}p{1.15cm}
    >{\centering\arraybackslash}p{1.15cm}
    >{\centering\arraybackslash}p{1.15cm}
    >{\centering\arraybackslash}p{1.15cm}}
\toprule
\textbf{Method}
& \textbf{F1}
& \textbf{Acc.}
& \textbf{P}
& \textbf{R} \\
\midrule
SP
& 41.6 & 55.0 & \underline{75.0} & 46.0 \\
MMD-Agent
& 57.3 & 55.8 & 60.6 & 56.9 \\
\midrule
\multicolumn{5}{l}{\textit{Minimal Prompt}} \\
Direct
& 45.2 & 51.3 & 57.6 & 49.7 \\
CoT
& 45.9 & 50.6 & 48.7 & 47.6 \\
Wiki Context
& 45.6 & 51.0 & 59.9 & 49.5 \\
ICL
& 54.6 & 54.6 & 70.8 & 57.6 \\
CR
& 46.9 & 51.9 & 60.0 & 49.1 \\
\midrule
\multicolumn{5}{l}{\textit{Verification Prompt}} \\
Direct
& 63.4 & 64.3 & 64.3 & 66.2 \\
CoT
& 62.1 & 63.1 & 63.1 & 64.4 \\
Wiki Context
& 65.2 & 65.7 & 65.2 & 67.8 \\
ICL
& 61.2 & 60.8 & 68.2 & 64.5 \\
ICL + Wiki 
& 64.0 & 63.5 & 67.0 & 67.3 \\
CR
& 64.7 & 64.2 & 67.7 & 67.5 \\
CR + Wiki
& \underline{67.8} & \underline{66.9} & {69.4} & \underline{69.3} \\
\midrule
\textbf{VNL (Ours)}
& \textbf{73.2}
& \textbf{71.8}
& \textbf{75.1}
& \textbf{72.4} \\
\bottomrule
\end{tabularx}
\vspace{2pt}
\end{table}

\begin{table}[t]
\centering
\caption{Ablation study of Verification-Notebook Learning. Bold denotes
the best result.}
\label{tab:ablation}
\small
\setlength{\tabcolsep}{5pt}
\begin{tabularx}{\columnwidth}{
    >{\raggedright\arraybackslash}X
    >{\centering\arraybackslash}p{1.8cm}
    >{\centering\arraybackslash}p{1.5cm}}
\toprule
\textbf{Variant}
& \textbf{F1}
& \textbf{Acc.} \\
\midrule
w/o Editor
& 63.4 & 64.7 \\
w/o Multi-Branch Selection
& 69.7 & 67.5 \\
w/o Wiki Context
& 70.2 & 69.2 \\
w/o Structured Sections
& 70.5 & 69.7 \\
\midrule
\textbf{Full VNL}
& \textbf{73.2}
& \textbf{71.8} \\
\bottomrule
\end{tabularx}
\end{table}

\begin{figure}[t]
\centering
\includegraphics[width=\columnwidth]{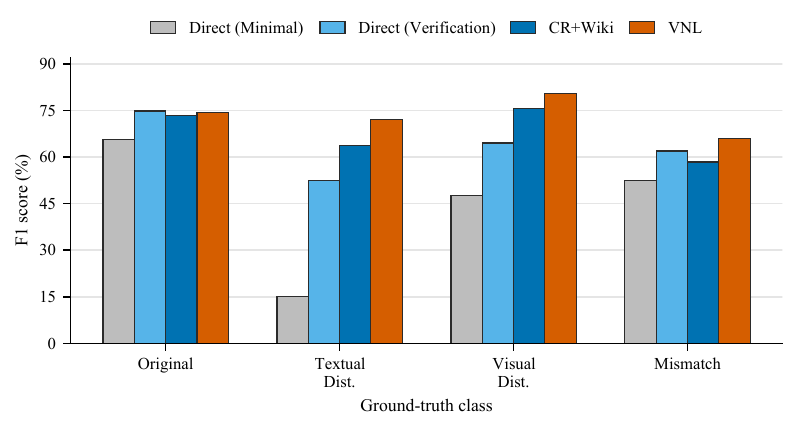}
\caption{Per-class F1 scores on the four-way source-attribution task.}
\label{fig:per_class_f1}
\end{figure}

\begin{figure*}[t]
\centering
\includegraphics[width=\textwidth]{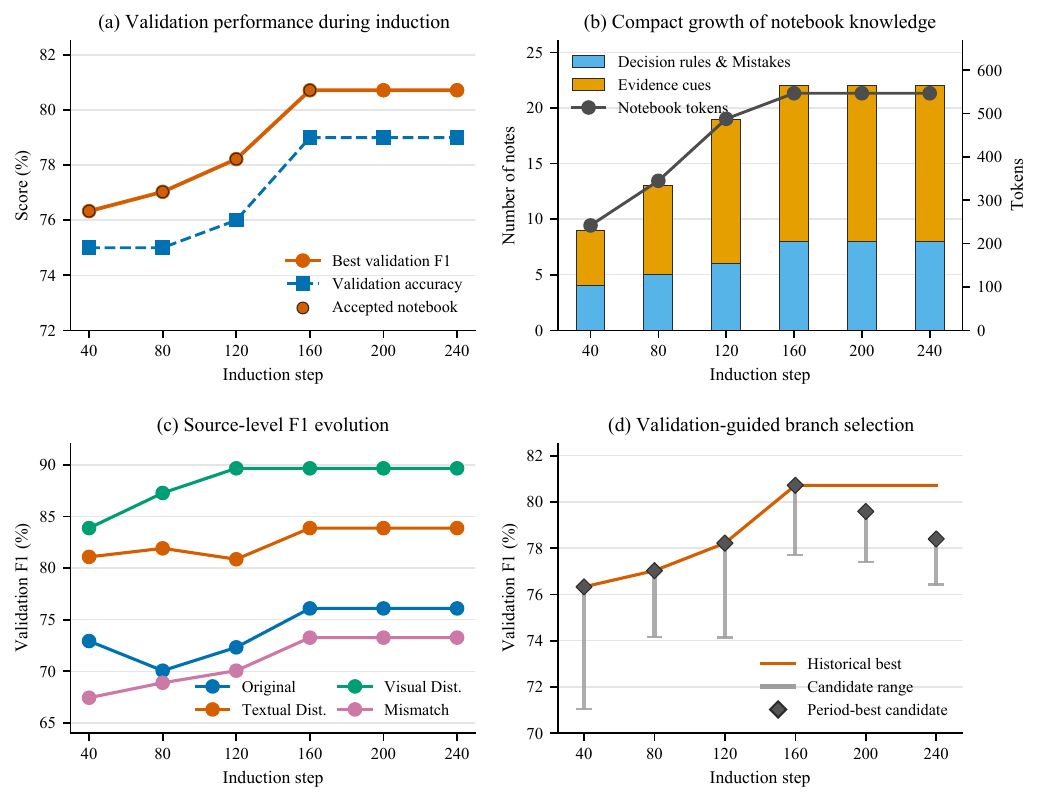}
\caption{Learning dynamics of one representative offline notebook induction run.}
\label{fig:notebook_evolution}
\end{figure*}

\subsection{Ablation Study}

\paragraph{Effect of notebook learning.}
The w/o Editor variant disables notebook induction and uses the Verifier without learned notebook updates. Its performance is close to that of Direct with the Verification Prompt, indicating that the verifier prompt already provides a strong task specification. Full VNL improves macro-F1 by 9.8 percentage points over this variant, showing that learned notebook knowledge accounts for most of the improvement. The notebook converts development feedback into reusable decision rules, observable evidence cues, and label-boundary guidance for source-aware verification.

\paragraph{Effect of validation-guided multi-branch selection.}
Removing validation-guided multi-branch selection reduces macro-F1 from 73.2\% to 69.7\%. A single induction path can still learn useful verification notes, but it is more likely to overemphasize incidental patterns in local batches or miss stronger combinations of notebook updates. Multi-branch selection addresses this issue by exploring several discrete update paths and selecting the branch that generalizes best to the held-out validation data. This result shows that notebook construction benefits from validation-guided search even when the LVLM parameters remain fixed.

\paragraph{Effect of external knowledge.}
Removing Wiki Context reduces macro-F1 from 73.2\% to 70.2\%, indicating that external evidence remains important for source-aware multimodal verification. Retrieved knowledge helps the Verifier check entity-level facts and resolve textual-veracity uncertainty when the post alone provides insufficient information. Nevertheless, the w/o Wiki Context variant still outperforms the other baselines. The learned notebook is therefore not simply a channel for external evidence. Even without retrieved context, it provides reusable decision rules and evidence-prioritization guidance that improve attribution among textual distortion, visual distortion, and mismatch.

\paragraph{Effect of structured notebook sections.}
Removing the structured sections reduces macro-F1 from 73.2\% to 70.5\%, showing that the organization of external knowledge affects how effectively it guides inference. Separating Decision Rules \& Mistakes to Avoid from Observable Evidence Cues clarifies the function of each note: some notes specify how to choose among source labels, whereas others identify concrete textual, visual, or cross-modal signals. With an unstructured collection, the Verifier must infer the function of each note during inference, which weakens the procedural role of the notebook.

\begin{figure*}[t]
    \centering
\includegraphics[
  width=\textwidth,
  trim=4pt 4pt 4pt 4pt,  % 左 下 右 上，按实际微调
  clip
]{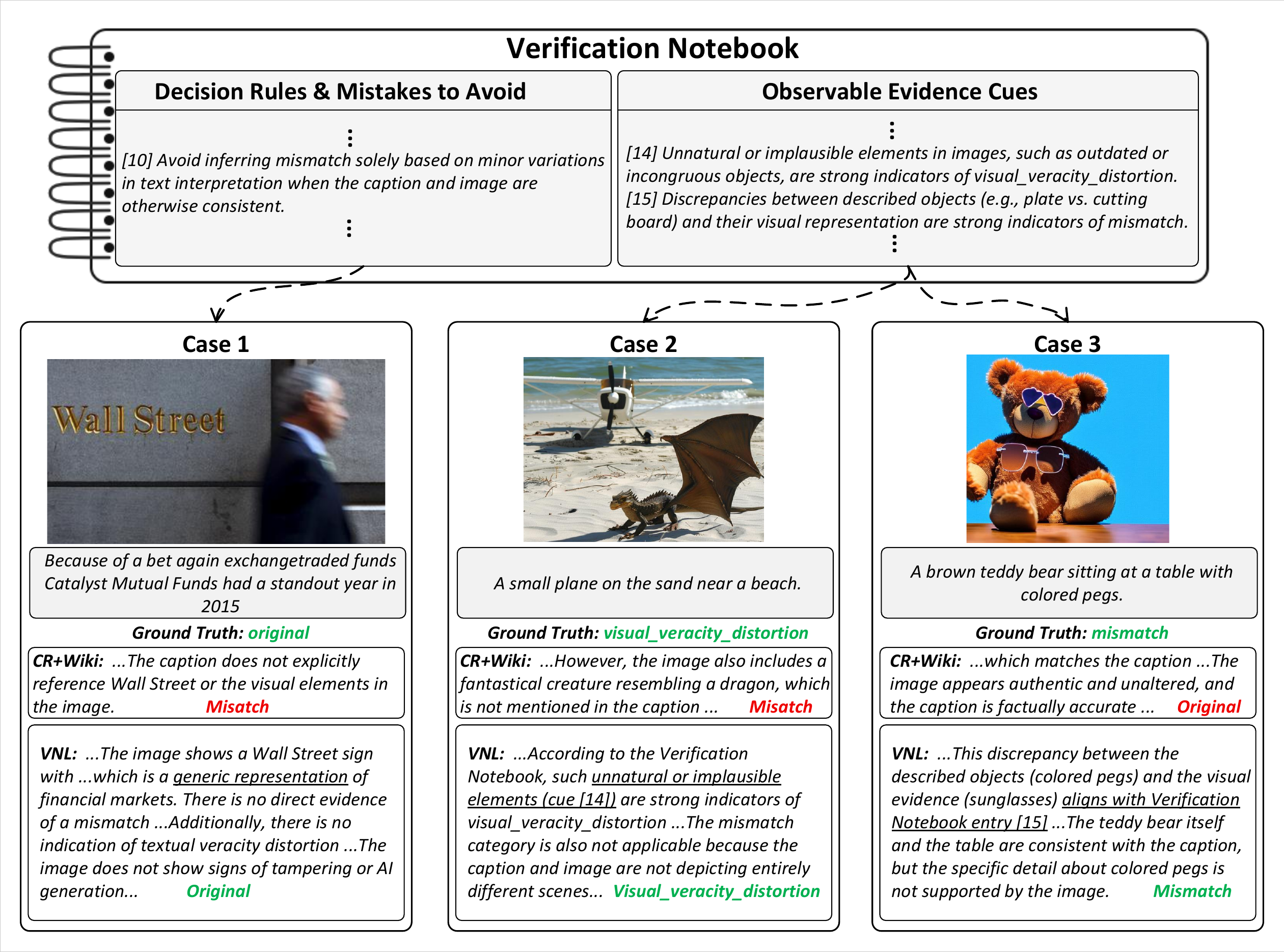}
    \caption{Case study of how a Verification Notebook guides different source-attribution decisions.}
    \label{fig:case_study}
\end{figure*}

\subsection{Per-Class Analysis}

Figure~\ref{fig:per_class_f1} compares the per-class F1 scores of representative methods. VNL achieves the strongest overall performance across source classes, particularly on the three misinformation classes. Its performance on the original class is comparable to that of the strongest direct baseline, while it achieves the highest F1 scores for textual distortion, visual distortion, and mismatch. This result is important because source-aware verification requires the model not only to identify reliable posts but also to attribute misleading posts to the correct source of distortion.

VNL consistently outperforms CR + Wiki across all three misinformation classes, with particularly large gains on textual distortion and mismatch. These improvements cannot be explained by access to retrieved evidence or case references alone. The notebook provides reusable guidance for source attribution, such as checking caption factuality before assigning mismatch and distinguishing visual manipulation from image-caption inconsistency. This guidance helps the Verifier avoid basing its decision solely on the most salient signal in each example.

VNL also produces a more balanced per-class performance profile. The difference between its strongest and weakest classes is smaller than that of CR + Wiki and substantially smaller than that of Direct with the Verification Prompt. Learned notebook guidance therefore improves both aggregate macro-F1 and the consistency of source attribution across different misinformation types.

\subsection{Learning Dynamics of Offline Notebook Induction}

We examine a representative induction trajectory to illustrate how the Verification Notebook is constructed before deployment. We track validation performance, notebook size, per-class performance, and branch selection at different induction checkpoints. This trajectory is illustrative rather than universal; it shows how VNL accumulates candidate verification knowledge offline and uses held-out validation feedback to select the notebook for inference.

Figure~\ref{fig:notebook_evolution}(a) shows the validation trajectory. Performance improves during the early stages of induction, and the best validation macro-F1 is reached after 160 development examples. Candidates produced at later checkpoints do not exceed this score, so the previous best notebook is retained. This trajectory illustrates the role of validation-based selection: notebook updates are retained only when they match or improve held-out performance, rather than simply because additional development examples have been processed.

Figure~\ref{fig:notebook_evolution}(b) shows that the notebook remains compact as its performance improves. At the selected checkpoint, it contains 22 notes and 547 tokens. Both notebook sections grow during induction. The decision-rule section records verification procedures and common mistakes, whereas the evidence-cue section records reusable textual, visual, and cross-modal signals. The resulting notebook is therefore a concise verification procedure rather than a memory of individual development examples.

Figure~\ref{fig:notebook_evolution}(c) presents the corresponding class-level trajectories. The improvements are not confined to a single label but occur across several source-attribution boundaries, including textual distortion, visual distortion, and mismatch. These results indicate that the notebook supports more consistent label-boundary decisions, such as separating false textual claims from cross-modal mismatch and distinguishing visual manipulation from image-caption inconsistency.

Figure~\ref{fig:notebook_evolution}(d) illustrates validation-guided branch selection. During each selection period, VNL evaluates several candidate notebooks and compares the strongest candidate with the historical best. Candidates from the early periods improve the selected notebook, whereas those from later periods fall below the historical best and are rejected. This procedure limits the effect of noisy or batch-specific updates and favors notebook changes that generalize beyond the local induction batches.

Figure~\ref{fig:case_study} presents three examples showing how the same learned Verification Notebook guides different source-attribution decisions. The cases cover three common decision boundaries: symbolic image-caption consistency, visual veracity distortion, and object-level mismatch.

\subsection{Case Study}

In the first case, the caption concerns Catalyst Mutual Funds, while the image shows a Wall Street sign. CR + Wiki predicts mismatch because the image does not directly depict the specific subject mentioned in the caption. VNL instead predicts original. The notebook guides the Verifier to interpret the Wall Street sign as a generic but relevant financial context rather than as evidence of contradiction. This rule prevents limited visual specificity from being mistaken for cross-modal mismatch.

The second case concerns the distinction between mismatch and visual veracity distortion. The caption states that a small plane is on the sand near a beach, which is consistent with the main content of the image. However, the image also contains an implausible dragon-like creature. CR + Wiki treats this unmentioned object as an image-caption mismatch. VNL assigns the example to visual veracity distortion because the error lies in the visual content itself rather than in a mismatch between unrelated textual and visual contexts. This decision follows the notebook cue that implausible or fabricated objects provide evidence of visual distortion.

The third case demonstrates the importance of examining local object-level evidence. The caption describes a brown teddy bear sitting at a table with colored pegs, whereas the image shows the teddy bear wearing sunglasses instead of the described pegs. CR + Wiki predicts original because the overall scene is broadly consistent with the caption. VNL predicts mismatch after checking whether the specific object described in the caption is present in the image. The notebook thus directs the Verifier beyond coarse scene similarity toward the object-level relation that determines the correct label.

Together, these cases show that VNL improves source attribution by refining the decision criteria used by the Verifier rather than by making the model uniformly more skeptical. The notebook helps the Verifier accept symbolic images when they provide an appropriate context, identify implausible visual content as visual distortion, and detect object-level contradictions even when the broader scene remains related. Verification-Notebook Learning thereby converts development experience into reusable rules and evidence cues that guide inference toward the correct source of misinformation.

\section{Conclusion}

We presented Verification-Notebook Learning, a non-parametric framework for source-aware multimodal misinformation detection with frozen LVLMs. Instead of updating model parameters or retaining individual cases, VNL learns an external verification procedure. It distills verification experience into a compact notebook of decision principles, evidence cues, and recurring pitfalls, which remains fixed and guides subsequent inference. This design provides an inspectable and persistent basis for evidence checking and source attribution while remaining compatible with black-box LVLMs. The experimental results and analyses show that the learned notebook outperforms strong inference-based alternatives by supporting more consistent verification decisions rather than memorizing individual cases. These findings demonstrate that reusable external procedures provide a practical form of learning for LVLM-based multimodal verification.

\bibliography{VNL}

\end{document}